\documentclass[letterpaper, 10 pt, conference]{ieeeconf}  
\IEEEoverridecommandlockouts
\usepackage{cite}
\usepackage{balance}
\usepackage{amsmath,amssymb,amsfonts}
\usepackage{subfigure}
\usepackage{graphicx} 
\usepackage{textcomp}
\usepackage{multirow} 
\usepackage{xcolor}
\usepackage{float}
\usepackage{tabularray}
\usepackage{hyperref}
\usepackage{array}
\usepackage{booktabs}
\usepackage{verbatim} 
\usepackage{makecell} 
\usepackage{booktabs} 
\usepackage{algorithm}
\usepackage{algpseudocode}
\makeatletter

\newcommand{\Rmnum}[1]{\expandafter\@slowromancap\romannumeral #1@}

\makeatother

\def\BibTeX{{\rm B\kern-.05em{\sc i\kern-.025em b}\kern-.08em
    T\kern-.1667em\lower.7ex\hbox{E}\kern-.125emX}}

\title{\LARGE \bf Sim4EndoR: A Reinforcement Learning Centered Simulation Platform for Task Automation of Endovascular Robotics}
\author{Tianliang Yao$^{1}$, Madaoji Ban$^{2}$, Bo Lu$^{3}$, Zhiqiang Pei$^{4}$, Peng Qi$^{1, 5, *}$
\thanks{This work has been submitted to the IEEE for possible publication. Copyright may be transferred without notice, after which this version may no longer be accessible.}
\thanks{The authors would like to thank Dr. Jingwei Song from the United Imaging Research Institute of Intelligent Imaging for providing technical support in experiments.}
\thanks{This work is supported by the National Key Research and Development Program of China under Grant No. 2023YFB4705200, the National Natural Science Foundation of China under Grant No. 62273257, and the Open Project Fund of State Key Laboratory of Cardiovascular Diseases No.2024SKL-TJ002. \emph{(*Corresponding Author: Peng Qi)}.}
\thanks{$^{1}$Department of Control Science and Engineering, College of Electronics and Information Engineering, and Shanghai Institute of Intelligent Science and Technology, Tongji University, Shanghai 200092, China;}%
\thanks{$^{2}$Department of Electrical and Electronic Engineering, Faculty of Engineering, The University of Hong Kong,  Hong Kong 999077, China;}
\thanks{$^{3}$Robotics and Microsystems Center, School of Mechanical and Electric Engineering, Soochow University, Suzhou, Jiangsu, China;}
\thanks{$^{4}$School of Oriental Pan-Vascular Devices Innovation College, University of Shanghai for Science and Technology, 516 Jungong Road, Shanghai 200093, China;}%
\thanks{$^{5}$State Key Laboratory of Cardiovascular Diseases and Medical Innovation Center, Shanghai East Hospital, School of Medicine, Tongji University 200092, Shanghai, China.}
}

\begin{document}

\maketitle 
\pagestyle{empty}  
\thispagestyle{empty} 

\begin{abstract}
Robotic-assisted percutaneous coronary intervention (PCI) holds considerable promise for elevating precision and safety in cardiovascular procedures. Nevertheless, current systems heavily depend on human operators, resulting in variability and the potential for human error. To tackle these challenges, Sim4EndoR, an innovative reinforcement learning (RL) based simulation environment, is first introduced to bolster task-level autonomy in PCI. This platform offers a comprehensive and risk-free environment for the development, evaluation, and refinement of potential autonomous systems, enhancing data collection efficiency and minimizing the need for costly hardware trials. A notable aspect of the groundbreaking Sim4EndoR is its reward function, which takes into account the anatomical constraints of the vascular environment, utilizing the geometric characteristics of vessels to steer the learning process. By seamlessly integrating advanced physical simulations with neural network-driven policy learning, Sim4EndoR fosters efficient sim-to-real translation, paving the way for safer, more consistent robotic interventions in clinical practice, ultimately improving patient outcomes.
\end{abstract}

\section{Introduction}
Robotic-assisted interventions in percutaneous coronary intervention (PCI) have been developed to enhance precision and control during cardiovascular procedures \cite{hoole2020recent}. These systems aim to address challenges including operator fatigue, radiation exposure, and limitations in manual dexterity  \cite{werner2018complex,yao2023enhancing}. However, current robotic systems are still fully reliant on the operator, resulting in time-consuming manual operations. To a certain extent, an improvement in the level of automation within these systems will enhance the intervention procedure and the physician's operative effect. For instance, Siemens Healthineers' Corindus, with its automated robotic movements in the technIQ Series designed for the CorPath$^\circledR$ GRX system, provides operators with new possibilities to advance and facilitate treatments \cite{Endovascular_Robotics}. Consequently, incorporating a certain degree of autonomous operation or intelligent assistance, particularly for aiding humans in navigation tasks that involve repetitive operations or complex lesions, is beneficial in future robotic-assisted PCI procedures.


Task-level autonomy, a pivotal milestone within the five-tier automation framework of medical robotics, entails robots autonomously executing specific tasks with minimal human intervention \cite{yang2017medical}. To attain this level, robots must be capable of interpreting clinical contexts and making decisions akin to those of experienced cardiologists. Nevertheless, the pursuit of high-level autonomy in surgical robotics is hampered by the necessity for robust, real-time decision-making amidst the intricacies of human anatomy and dynamic surgical environments \cite{zhong2023integrated}. Reinforcement learning (RL) \cite{omisore2021novel} presents a promising avenue, empowering robots to learn optimal strategies through iterative trial and error in simulated settings. This methodology holds significant potential for enhancing surgical robot autonomy in complex procedures such as PCI \cite{fiorini2022concepts}. However, the development and validation of such systems in real-world contexts pose challenges owing to the time, resources, and risks associated with the process.


\begin{figure*}[!htbp]
\centering
\includegraphics[width=0.93\textwidth]{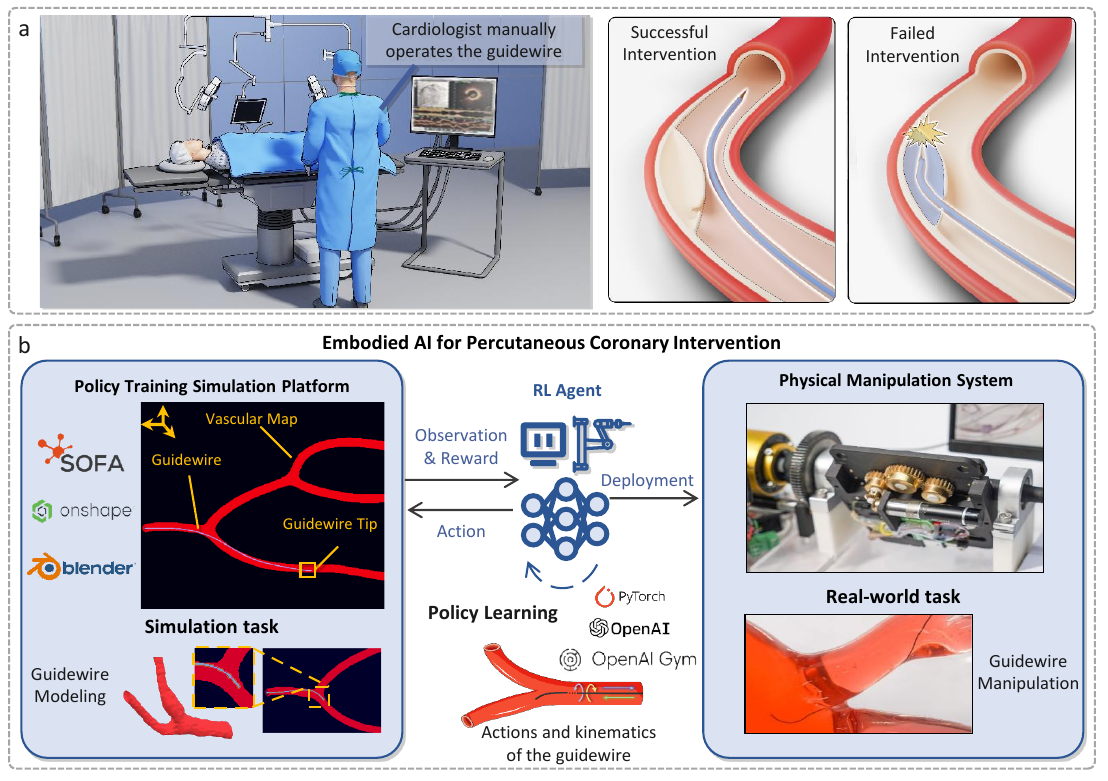}
\caption{Framework of Sim4EndoR for achieving embodied AI in PCI procedures: (a) Traditional interventional cardiologists require a prolonged learning curve to attain proficiency in intricate guidewire manipulation techniques. Additionally, engaging in numerous interventional procedures over an extended period can cause fatigue and radiation-induced illnesses among cardiologists. This, in turn, may lead to unsuccessful procedures or associated complications in complex cases, such as vessel perforation. (b) The proposed Sim4EndoR encompasses a simulation platform dedicated to policy training. Within this platform, the RL agent simulates the actions and kinematics of the guidewire, learning optimal manipulation strategies through observation and reward mechanisms. Furthermore, Sim4EndoR incorporates a robotic guidewire delivery system designed for real-world applications. Policy deployment is facilitated by the physical manipulation system, enabling precise guidewire manipulation that ultimately contributes to successful interventions.}
\label{fig:OverallFramework}
\end{figure*}

While platforms such as SurRoL \cite{xu2021surrol, long2022integrating, huang2023guided}, Surgical Gym \cite{schmidgall2024surgical}, and Orbit \cite{yu2024orbit} have facilitated the advancement of da Vinci$^\circledR$ surgical robotics, the field of endovascular robotics lacks similarly exhaustive simulation and testing environments \cite{robertshaw2023artificial}. Existing platforms, including CASOG and the Model-Based Offline RL by Li et al. \cite{li2023casog,li2024model}, are confined to two-dimensional (2D) planes and fail to replicate the three-dimensional (3D) complexities inherent in PCI procedures, thereby inadequately addressing clinical requirements. The primary limitation of 2D environments lies in their inability to accurately mimic intricate 3D anatomical structures. Consequently, the challenges of navigating tortuous blood vessels \cite{tamadon2023semiautonomous} and managing bifurcations within the vasculature are significantly oversimplified \cite{jingwei2024vascularpilot3d}. To address this crucial shortcoming, we introduce Sim4EndoR, a reinforcement learning-focused 3D simulation environment tailored to propel task-level autonomy in endovascular robotics. Sim4EndoR aims to bridge the divide between current robotic capabilities and the heightened levels of autonomy demanded by complex, high-stakes cardiovascular intervention procedures.


Figure \ref{fig:OverallFramework} depicts the conceptual framework of an embodied AI system aimed at achieving task-level autonomy in PCI, a capability that remains unfulfilled by current simulation systems. Within this framework, the RL agent formulates policies within the simulation environment, which are subsequently transferred to real-world robotic systems for guidewire manipulation. Through iterative policy learning, the agent continually refines its strategies, guaranteeing efficient performance in clinical applications. By merging advanced RL algorithms with highly realistic simulations of PCI, Sim4EndoR establishes an intelligent and autonomous platform for the development, evaluation, and optimization of autonomous robotic-assisted cardiovascular interventions, thereby expanding the horizons of simulation-based PCI training and robotics \cite{scarponi2024zero, robertshaw2024autonomous}.


The primary contributions of this work are outlined as follows:
\begin{itemize}
    \item \textbf{Reinforcement Learning Framework:} A novel RL framework is introduced, specifically tailored to address the inherent challenges posed by 3D environments and continuous action spaces in PCI procedures.
    \item \textbf{Reward Function Design:} A reward function is proposed that defines the distance between the real-time position and the target on the \textit{Riemannian }manifold.
    \item \textbf{Real-World Policy Deployment:} We deployed trained policies on endovascular robotics platforms, demonstrating that Sim4EndoR, with its rich physical interactions incorporated, excels in real-world transferability.
\end{itemize}

\section{Methodology}

\subsection{System Overview}
The proposed Sim4EndoR, as depicted in Fig. \ref{fig:OverallFramework}(b), seamlessly integrates advanced physical simulation with real-time rendering, thereby crafting a highly realistic training environment exclusively tailored for PCI procedures. A pivotal feature of Sim4EndoR lies in its meticulous modeling of the guidewire, which is central to PCI procedures. This modeling accurately captures the guidewire's flexibility and adaptability, facilitated by efficient collision detection and kinematic mapping mechanisms. These functionalities are realized utilizing the Simulation Open Framework Architecture (SOFA) \cite{faure2012sofa}, a robust platform for real-time physical simulation. To further bolster realism, Sim4EndoR incorporates a neural network for real-time processing of the guidewire's operational state, facilitating intelligent policy learning. The RL agent leverages this data to model the guidewire's actions and kinematics, progressively learning optimal manipulation strategies through observation and reward-based mechanisms.

A notable aspect of Sim4EndoR is its capability for sim-to-real transfer. The policies honed within the simulation environment are designed to seamlessly transition to real-world robotic systems. This is achieved by meticulously calibrating simulation parameters to closely mirror real-world conditions, encompassing the mechanical characteristics of both the guidewire and vascular tissues. Consequently, once trained, the RL agent's policies can be deployed in robotic guidewire delivery systems for real-world applications.



\subsection{Simulation Environment}
Sim4EndoR is designed to facilitate seamless human-robot interaction via an intuitive user interface that offers real-time scene rendering. This interface facilitates the transmission of specific action encoding vectors, which are tailored to align with the motion patterns of PCI procedure guidewires. This capability simplifies the invocation of parameters during simulations.

The virtual assets within Sim4EndoR are meticulously crafted using a combination of advanced software tools. The SOFA physics engine (v23.06) provides a robust foundation for simulating intricate interactions between the guidewire and anatomical structures, leveraging finite element methods to model deformable vascular tissues and collision detection for guidewire dynamics. Blender is employed to create detailed 3D models, including patient-specific coronary artery geometries derived from 3D CTA imaging and guidewire/catheter models reflecting real-world specifications. These models are then imported into Onshape, a cloud-based parametric CAD software, where they are configured and calibrated. In Onshape, precise coordinate systems are established to align the vascular models and guidewire models with the robotic reference frame. Properties such as vessel curvature, guidewire stiffness, and surface boundaries are parametrically defined and fine-tuned to ensure the simulations accurately replicate the complexities of real-world PCI procedures.



\subsection{Reinforcement Learning Framework for PCI}

To achieve task-level navigation autonomy in endovascular robotics, we formulate the decision-making problem as a Markov Decision Process (MDP) \cite{li2024model}, defined by the tuple $\mathcal{M} = \langle \mathcal{S}, \mathcal{A}, \mathcal{P}, \mathcal{R}, \gamma \rangle$, where:

\begin{itemize}
    \item $\mathcal{S}$ represents the state space, which includes the robot's configuration, the guidewire's position and orientation, the vascular anatomy, and relevant physiological parameters encompass vascular curvature, which quantifies the local bending of the vessel, and the spatial structure, representing the 3D vascular topology derived from preoperative imaging..
    \item $\mathcal{A}$ denotes the action space, defined by the robot's control inputs, specifically the translational and rotational movements of the guidewire.
    \item $\mathcal{P}(s'|s,a)$ is the state transition probability function, representing the likelihood of transitioning from state $s$ to state $s'$ after executing action $a$.
    \item $\mathcal{R}(s,a)$ is the reward function, which assigns scalar rewards to state-action pairs to encourage successful and safe task execution.
    \item $\gamma$ is the discount factor, which balances the importance of immediate versus future rewards.
\end{itemize}

\subsection{State and Action Space Representation}

At each discrete time step $t$, the state of the system, denoted as $s_t \in \mathcal{S}$ ($\mathcal{S}$ is the state space), is represented by a vector that captures the current configuration of the guidewire's tip as well as its dynamic properties. Specifically, the state vector $s_t$ is defined as follows:

\begin{equation}
    s_t = \left[ \mathbf{p}_t, \mathbf{v}_t \right]^\top
\end{equation}

where:
\begin{itemize}
    \item $\mathbf{p}_t \in \mathbb{R}^3$ represents the 3D position of the guidewire tip in the vessel at time step $t$,  expressed in Cartesian coordinates.
    \item $\mathbf{v}_t \in \mathbb{R}^3$ denotes the velocity vector of the guidewire tip, reflecting the instantaneous rate of change of the tip's position.
\end{itemize}

The action $a_t$ at each time step $t$ defines the set of permissible control inputs that the robotic system can apply to the guidewire. The action $a_t \in \mathcal{A}$ ($\mathcal{A}$ is the action space) is a vector that specifies the incremental changes in the guidewire's configuration:

\begin{equation}
    a_t = \left[ \Delta d_t, \Delta \theta_t \right]^\top
\end{equation}

where:
\begin{itemize}
    \item $\Delta d_t \in \mathbb{R}$ represents the incremental distance by which the guidewire is either inserted into or retracted from the vessel. This scalar value controls the transnational movement of the guidewire tip.
    \item $\Delta \theta_t \in \mathbb{R}$ denotes the incremental rotational angle applied to the guidewire around its longitudinal axis. This angular adjustment allows the guidewire to rotate, enabling it to navigate through curved or branching sections of the vascular tree.
\end{itemize}

\subsection{Reward Function Design}

Traditional Euclidean distance metrics fall short of capturing the intricacies of vascular environments \cite{van2024geodesic}, where the trajectory of a guidewire is significantly impacted by the curvature and topology of the vessels. To overcome these limitations, we have defined a reward function $\mathcal{R}(s_t, a_t)$ that is grounded in curvature distance on the vascular manifold $\mathcal{M}$.


The reward function at time $t$ is:

\begin{equation}
\mathcal{R}(s_t, a_t) = - d_{\mathcal{M}}(\mathbf{p}_{t-1}, \mathbf{p}_{\text{goal}}),
\end{equation}

\noindent wherein the curvature  distance $d_{\mathcal{M}}(s_t, \mathbf{p}_{\text{goal}})$ is defined by:

\begin{equation}
d_{\mathcal{M}}(\mathbf{p}_{t}, \mathbf{p}_{\text{goal}}) = \int_{\tau_0}^{\tau_{goal}}\sqrt{\left(\dot{\gamma}_x(\tau)\right)^2+\left(\dot{\gamma}_y(\tau)\right)^2+\left(\dot{\gamma}_z(\tau)\right)^2}d\tau,
\end{equation}

\noindent The curvature distance is defined along a smooth curve \(\gamma(\tau): [\tau_0, \tau_{\text{goal}}] \rightarrow \mathcal{M}\), where \(\mathcal{M}\) is a Riemannian manifold representing the vascular geometry, and \(\tau\) parameterizes the curve (e.g., as arc length). Here, \(\tau_0\) is the initial parameter at the guidewire’s current tip position \(\mathbf{p}_t\), and \(\tau_{\text{goal}}\) is the terminal parameter at the target position \(\mathbf{p}_{\text{goal}}\), with \(\gamma(\tau_0) = \mathbf{p}_t\) and \(\gamma(\tau_{\text{goal}}) = \mathbf{p}_{\text{goal}}\). The components \(\gamma_x\), \(\gamma_y\), and \(\gamma_z\) denote the projections of \(\gamma\) onto the \(x\)-, \(y\)-, and \(z\)-axes, with \(\dot{\gamma}_x\), \(\dot{\gamma}_y\), and \(\dot{\gamma}_z\) as their first-order derivatives with respect to \(\tau\), reflecting the curve’s tangent. In practice, this distance is approximated using Dijkstra’s algorithm on a discretized graph representation of \(\mathcal{M}\). The manifold is modeled as a network of nodes (sampled vascular points) and edges, with edge weights calibrated to local geometric properties (e.g., curvature and tortuosity), providing an efficient approximation of the curvature-aware path from \(\mathbf{p}_t\) to \(\mathbf{p}_{\text{goal}}\).

The reward function is designed to encourage the agent's progress toward the target and is defined as follows:

\begin{itemize}
\item \textbf{Reduction Reward:} The reward depends on the reduction in curvature distance to the target. A positive reward is given if the distance decreases from the previous step:

\begin{equation}
\mathcal{R}_{\text{dist}} = \frac{D_{\text{last}} - D_{\text{current}}}{100},
\end{equation}

where $D_{\text{last}} = d_{\mathcal{M}}(\mathbf{p}_{t-1}, \mathbf{p}_{\text{goal}})$ is the curvature distance to the target from the previous position, and $D_{\text{current}} = d_{\mathcal{M}}(\mathbf{p}_t, \mathbf{p}_{\text{goal}})$ is the current curvature distance to the target.

\item \textbf{Goal Achievement Bonus:} An additional reward is assigned when the guidewire reaches a proximity threshold of $1 mm$ to the target position:

\begin{equation}
    \mathcal{R}_{\text{goal}} = 100, \quad \text{if } D_{\text{current}} \leq 1.
\end{equation}

This bonus incentivizes the agent to prioritize reaching the target location.

\item \textbf{Failure Penalty:} A penalty is imposed if the guidewire encounters a failure state, including exceeding operational limits and colliding with vessel walls: $\mathcal{R}_{\text{fail}} = -100.$

\end{itemize}

The reward function is designed to encourage the agent to navigate effectively through the vascular network by balancing exploration with safety, and the penalty discourages unsafe maneuvers and reinforces adherence to the anatomical constraints of the vascular structure. By incentivizing the reduction of curvature distance to the target, the reward function aligns the guidewire's actions with the clinical objective of accurately navigating to specific regions, such as stenosed artery segments.

\section{Experiments and Results}
\subsection{Implementation Details}

Sim4EndoR was implemented using a combination of high-fidelity physics engines, including SOFA 23.06.00 \cite{faure2012sofa} and Blender \cite{blender_org}, along with medical imaging data to construct realistic and detailed coronary artery models. The simulation environment was designed to be highly modular and flexible, allowing easy customization and extension. The modularity facilitates the integration of new anatomical models, surgical tools, and physiological conditions, enabling a wide range of experimental setups and research scenarios.

To ensure efficient training and evaluation of models, an experimental platform was deployed on a high-performance computing cluster equipped with an NVIDIA GeForce GTX1660 Ti GPU. The implementation leveraged  PyTorch 1.11.0 framework \cite{paszke2019pytorch}, with Python 3.8. RL algorithms were implemented using the Gym library developed by OpenAI \cite{brockman2016openai}. 

In the simulation environment, several key hyperparameters were carefully selected. Specifically, a target smoothing coefficient of $\tau=0.005$ was employed, with the target network updated at an interval of 1, and 10 test iterations conducted. The learning rate was set to $1 \times 10^{-4}$, and the discount factor $\gamma$ was chosen to be 0.99. The replay buffer was configured with a capacity of 1 million entries, and the mini-batch size for sampling was set to 100. To balance exploration and exploitation, exploration noise with a magnitude of 0.1 was introduced, with sampling performed every 2000 steps. The training progress was logged at intervals of 50 iterations, and the environment was rendered every 100 iterations. Training was conducted for a total of 100,000 steps. No pre-trained models were loaded during the training process, ensuring results derived from scratch.

The algorithms were trained and evaluated on two distinct vascular phantom models, highlighting the generalization ability of the algorithm. The Simplified Vascular Phantom is derived from the standard Vessel Model included in the SOFA platform and was used extensively in the simulations. The vascular phantom was obtained from phantoms and by a CT scanner from the United Imaging Research Institute of Intelligent Imaging. The vascular phantom includes a corresponding physical counterpart, which was utilized for the final Sim2Real deployment. 

\begin{figure}[!htbp]
\centering
\includegraphics[width=0.94\linewidth]{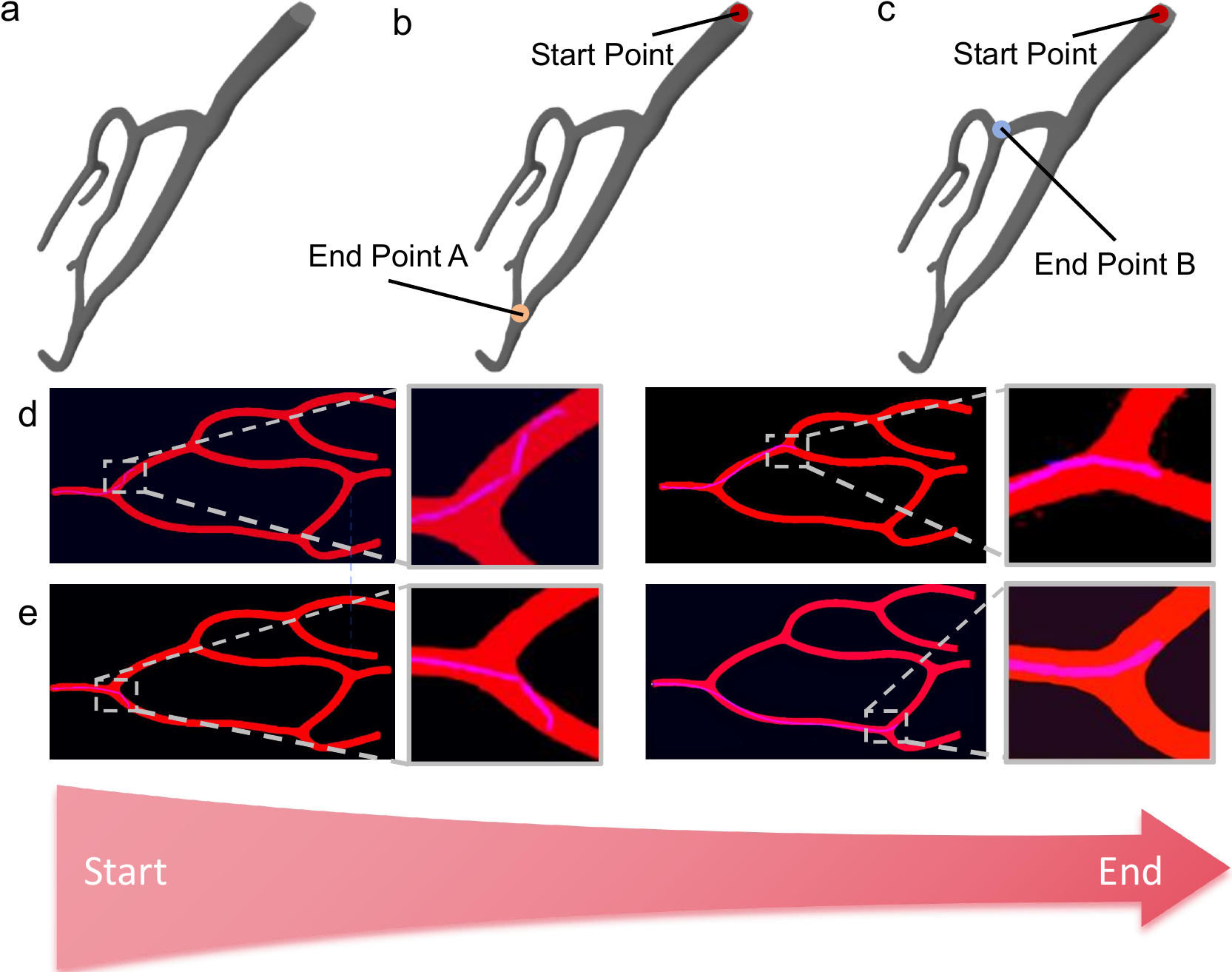}
\caption{Illustration is the guidewire navigation task within the Simplified Vascular Phantom. (a) The vascular network with bifurcation points. (b) Task A: Navigation to End Point A. (c) Task B: Navigation to End Point B. (d) and (e) show the guidewire reaching the designated End Points A and B, respectively, within the simulated environment.}
\label{fig:Task1}
\end{figure}

\subsection{Simulation Task Configuration and Validation}
The navigation tasks were meticulously designed to evaluate the accuracy and efficiency of guidewire navigation in simulation which is similar to the real-world intravascular procedure tasks. Two distinct models were utilized: the Simplified Vascular Phantom and the Complex Vascular Phantom. These models simulate varying anatomical scenarios to assess the performance of various RL algorithms.


\subsubsection{Simplified Vascular Phantom}
As depicted in Fig.~\ref{fig:Task1}, the Simplified Vascular Phantom represents a simplified vascular network featuring a bifurcation. The guidewire commences its journey from a predetermined Start Point, with the primary goal being to navigate it through the vascular pathways to reach one of the designated End Points (A or B). The task is deemed successful when the tip of the catheter is within 1 mm proximity of the selected End Point. This model functions as an initial testbed to evaluate the RL algorithms' capability in handling fundamental navigation tasks.




\begin{table*}[!htbp]
\centering
\caption{Performance Comparison of RL Algorithms Across Two Simulation Tasks on Simplified Vascular Phantom}
\begin{tabular}{c c c c c c c c}
\Xhline{2\arrayrulewidth}
\multirow{2}{*}{\textbf{Algorithm}} & \multicolumn{3}{c}{\textbf{Task A}} & \multicolumn{3}{c}{\textbf{Task B}} \\ 
\cmidrule(lr){2-4} \cmidrule(lr){5-7}
 & \textbf{Trials} & \textbf{Success Rate} & \textbf{\makecell{Average Completion \\ Time (s)}} & \textbf{Trials} & \textbf{Success Rate} & \textbf{\makecell{Average Completion \\ Time (s)}} \\ 
\Xhline{2\arrayrulewidth}
DDPG (manifold distance reward) & 10 & 80\% (8/10) & 5.27  & 10 & 80\% (8/10) & 7.81  \\
DDPG (euclidean distance reward) & 10 & 30\% (3/10)  & 5.39  & 10 & 10\% (1/10)  & 7.84  \\
PPO (manifold distance reward) & 10 & 70\% (7/10) & 5.43 & 10 & 80\% (8/10) & 8.02  \\
SAC (manifold distance reward) & 10 & 80\% (8/10) & 5.69 & 10 & 80\% (8/10) & 7.87 \\
\Xhline{2\arrayrulewidth}
\end{tabular}
\label{tab:simulation_results}
\vspace{1em}  
\footnotesize
\par Note: In DDPG (euclidean distance reward), 5 out of 10 random seeds did not converge in Task A, while 8 out of 10 did not converge in Task B.
\end{table*}


\subsubsection{Vascular Phantom}
\begin{figure}[!htbp]
\centering
\includegraphics[width=0.9\linewidth]{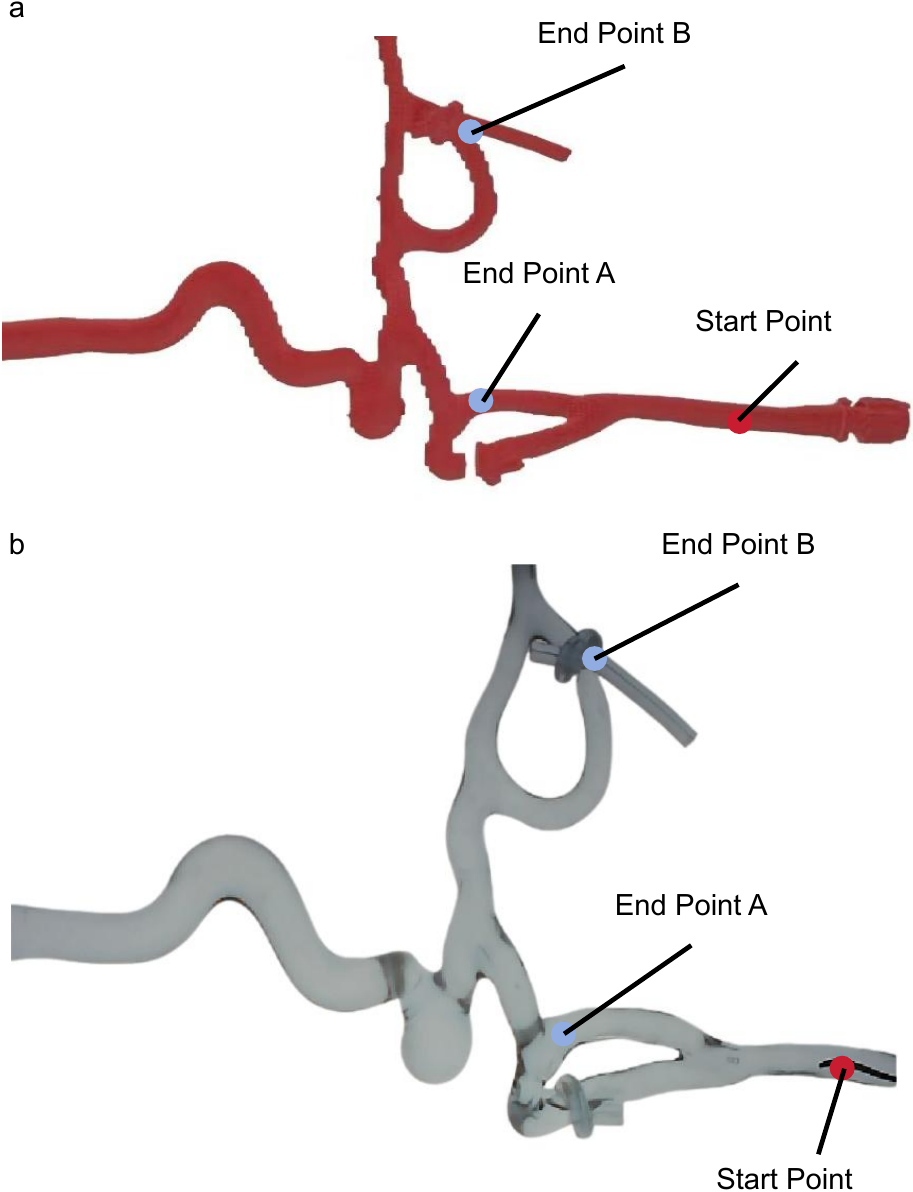}
\caption{The simulation of the Vascular Phantom, which corresponds to a real-world physical model for deployment, initiates navigation from the Start Point with the objective of accurately reaching one of the targeted End Points (A or B).}
\label{fig:VascularPhantom}
\end{figure}

The vascular phantom showcased in Fig. \ref{fig:VascularPhantom} incorporates a diverse array of both straight and curved anatomical structures. What distinguishes this model is its tangible counterpart, tailored exclusively for evaluating algorithm deployment in Sim2Real scenarios.

\subsubsection{Performance Evaluation}

To validate the efficacy of the RL algorithms, we performed a series of experiments across the two models. The algorithms were evaluated using two metrics: Success Rate and Completion Time. Success Rate signifies the reliability and consistency of the algorithms in achieving the target endpoints, while Completion Time assesses their efficiency by measuring the time taken to complete the navigation task.

In the Simplified Vascular Phantom, the performance of the three RL algorithms—Deep Deterministic Policy Gradient (DDPG), Proximal Policy Optimization (PPO), and Soft Actor-Critic (SAC)—was evaluated across two navigation tasks. As shown in Table~\ref{tab:simulation_results}, all algorithms demonstrated comparable success rates, with the highest at $80\%$. DDPG achieved the fastest average completion time for both tasks, followed by SAC and PPO, indicating a slight advantage in efficiency.




\begin{figure*}[!htbp]
\centering
\includegraphics[width=0.85\linewidth]{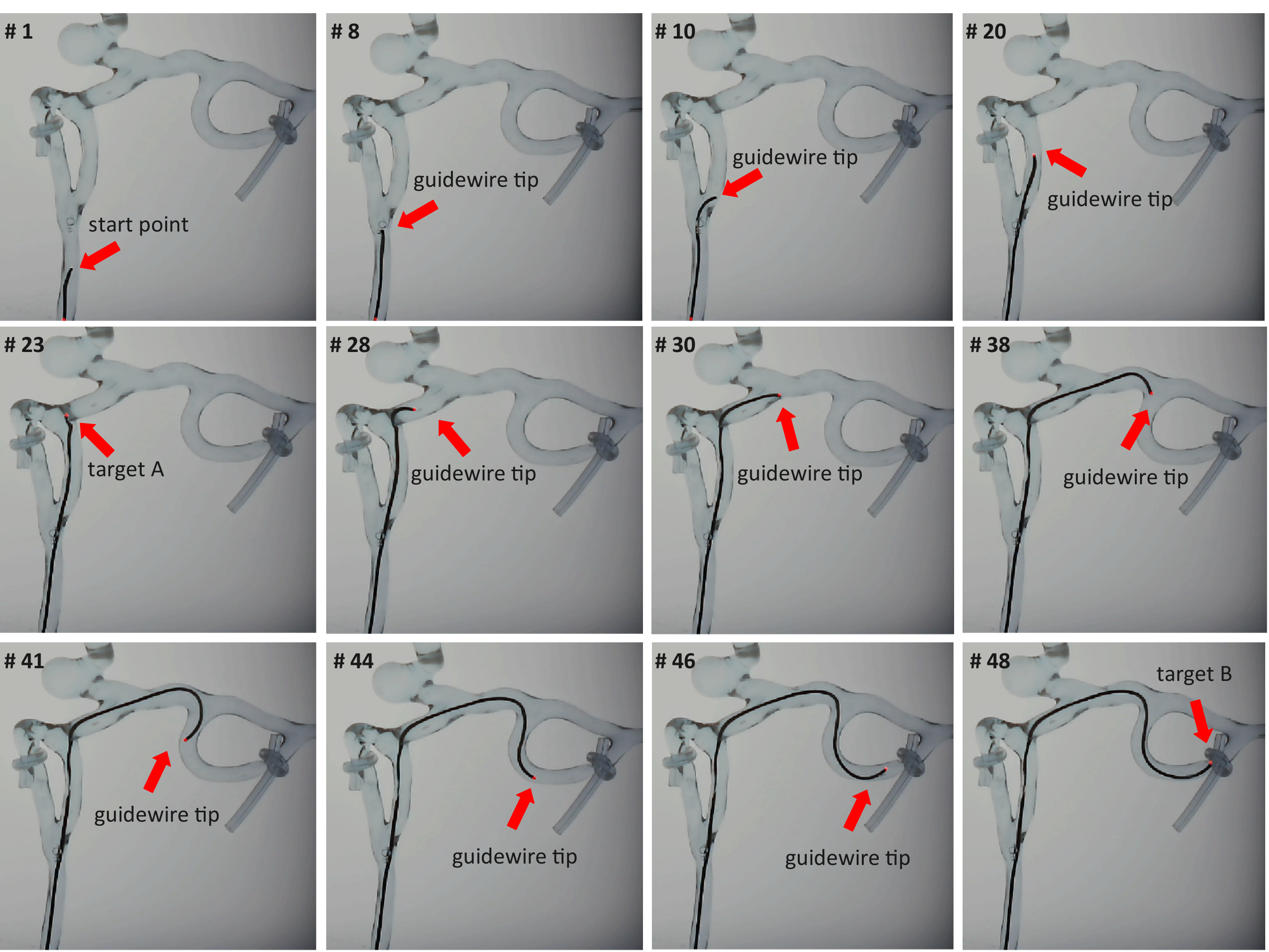}
\caption{The sequence of keyframes extracted from recorded videos demonstrates the autonomous navigation of the guidewire, which is achieved based on the proposed skill learning paradigm. Notably, the guidewire adeptly navigates through multiple vascular bifurcations to reach the designated target location, all without relying on any real-time position feedback.}
\label{fig:Deployment}
\end{figure*}

The simulation results on the vascular phantom, based on DDPG with an improved reward function, were also promising. For Task A, all 5 trials were successful, while for Task B, 4 out of 5 trials resulted in success. 

\subsection{Physical Robot Deployment}
To evaluate the effectiveness of the proposed navigation algorithms in a practical context, we deployed them on a physical robotic platform provided by the United Imaging Research Institute of Intelligent Imaging, which was purposely designed for intravascular procedures. The robotic hardware system possesses the capability to stably and precisely manipulate the guidewire.

The deployment encompassed two tasks, Task A and Task B, each representing varying levels of complexity. In Task A, all five trials were successful, showcasing the robustness of the algorithm in scenarios of lower complexity. For Task B, which posed a more intricate navigation challenge, the algorithm achieved success in three out of five trials, underscoring its strong performance while also highlighting areas requiring further optimization. Illustrated in Fig. \ref{fig:Deployment} are key frames from a successful deployment, demonstrating the robot's capability to execute the trained actions with ease, navigating smoothly through multiple vascular bifurcations and reaching the target point with precision.





\section{Discussion}
The utilization of the high-fidelity Sim4EndoR simulation environment effectively bridges the gap between theoretical RL frameworks and their practical implementation in robotic systems. Nevertheless, challenges in the Sim2Real transfer persist. Notably, the silicone-based vascular model exhibits physical properties that diverge from those of human tissue, resulting in discrepancies within the simulation. Additionally, the current physical modeling of the guidewire, particularly its tip, necessitates further refinement. Current simulation tools fall short in accurately replicating the intricate nonlinear behaviors of the guidewire, which are pivotal for successful navigation and manipulation within the vascular environment.

\section{Conclusion and Future Work}
This paper introduces Sim4EndoR, a simulation learning paradigm centered on RL for enhancing autonomy in PCI robotics. By incorporating state-of-the-art RL algorithms within a meticulously crafted simulation environment, Sim4EndoR effectively bridges the existing gap between robotic capabilities and the level of autonomy necessary for intricate surgical tasks. To the best of the authors' knowledge, this represents the first proposal for skill training based on a virtual simulation platform specifically designed for vascular interventional robots. Notably, the system has exhibited successful simulation-to-reality (Sim2Real) transfer, achieving success rates exceeding 70\% across diverse tasks in both simulated and real-world contexts.

Future research endeavors will concentrate on refining RL algorithms to conquer even more sophisticated and demanding surgical challenges \cite{iyengar2023sim2real}, particularly those involving dynamic anatomical fluctuations and patient-specific variabilities. Additionally, the development of more advanced simulation environments, especially those capable of accurately replicating interactions between soft tissues such as blood vessels and guidewires, will be crucial for fostering the creation of more robust simulation systems. With continued enhancements, Sim4EndoR holds the promise of becoming an invaluable asset to cardiologists, ultimately elevating the efficacy and outcomes of PCI procedures.




\bibliographystyle{ieeetr}
\balance
\bibliography{reference}
\end{document}